\pdfoutput=1

\documentclass[11pt]{article}

\usepackage{acl}

\usepackage{times}
\usepackage{latexsym}

\usepackage[T1]{fontenc}

\usepackage[utf8]{inputenc}

\usepackage{microtype}
\usepackage{times}
\usepackage{graphicx}
\usepackage{latexsym}
\usepackage{amsmath}
\usepackage{amsfonts}
\usepackage{amssymb}
\usepackage{mathtools}
\usepackage{multirow}
\usepackage{url}
\usepackage{adjustbox}
\usepackage{tabularx}
\usepackage{textcomp}
\usepackage{booktabs}
\usepackage[inline, shortlabels]{enumitem}

\usepackage{xcolor}

\title{Graph Refinement for Coreference Resolution}

\author{Lesly Miculicich \thanks{\hspace{2mm}Work done as a PhD student at EPFL/Idiap}  \\ Microsoft \\ \texttt{leslym@microsoft.com} 
        \And James Henderson \\ Idiap Research Institute  \\ \texttt{james.henderson@idiap.ch}}

\begin{document}
\maketitle
\begin{abstract}
The state-of-the-art models for coreference resolution are based on independent mention pair-wise decisions. We propose a modelling approach that learns coreference at the document-level and takes global decisions. For this purpose, we model coreference links in a graph structure where the nodes are tokens in the text, and the edges represent the relationship between them. Our model predicts the graph in a non-autoregressive manner, then iteratively refines it based on previous predictions, allowing global dependencies between decisions. The experimental results show improvements over various baselines, reinforcing the hypothesis that document-level information improves conference resolution. \end{abstract}

\section{Introduction} 
\label{sec:coreference:intro}


Current state-of-the-art (SOTA) solutions for coreference resolution such as \cite{toshniwal-etal-2020-learning, xu-choi-2020-revealing, wu-etal-2020-corefqa} formulate the problem in an end-to-end manner where the models jointly learn to detect mentions and link coreferent mentions. The objective is to predict the antecedent of each mention-span in a document, so the model performs pair-wise decisions of all mentions. After having the model predictions, related mentions are grouped into clusters. Under this scenario, each decision (i.e., whether two mentions are related to the same entity or not) is independent.
\citet{lee-etal-2018-higher} proposed an iterative method to update the representation of a mention with information of its probable antecedents. However, the final decisions are still made locally. 

\begin{figure}
	\center
	\includegraphics[width=1\linewidth]{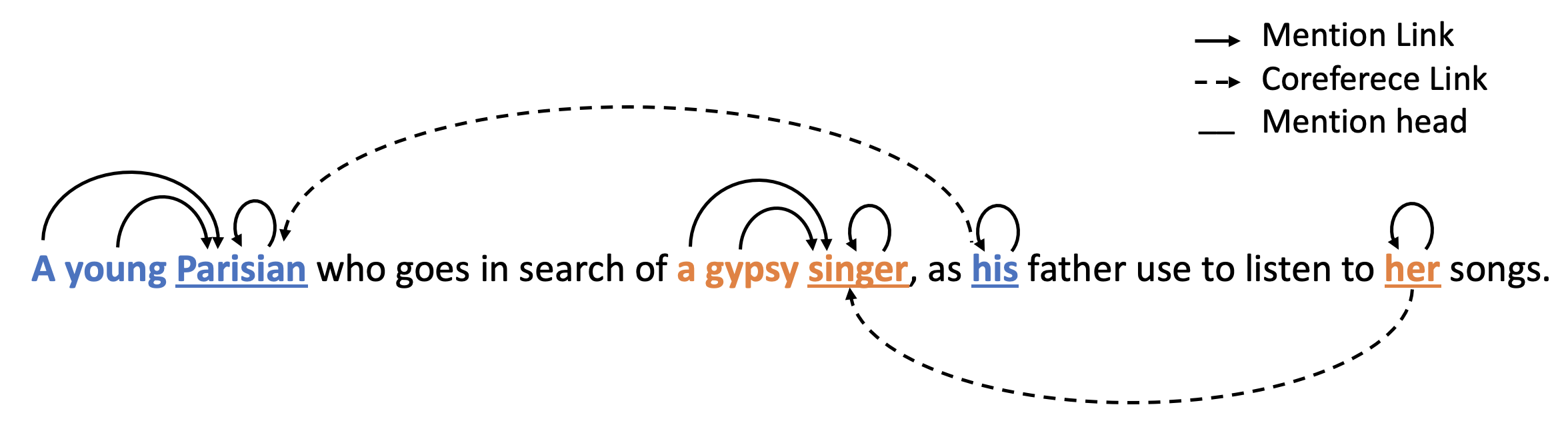}
	\caption{Example of a graph structure for coreference. Mention spans are shown in bold, and colors represent entity clusters. The mention heads are underlined.}
	\label{fig:coreference:exa}
\end{figure}

\begin{figure}
	\center
	\includegraphics[width=.8\linewidth]{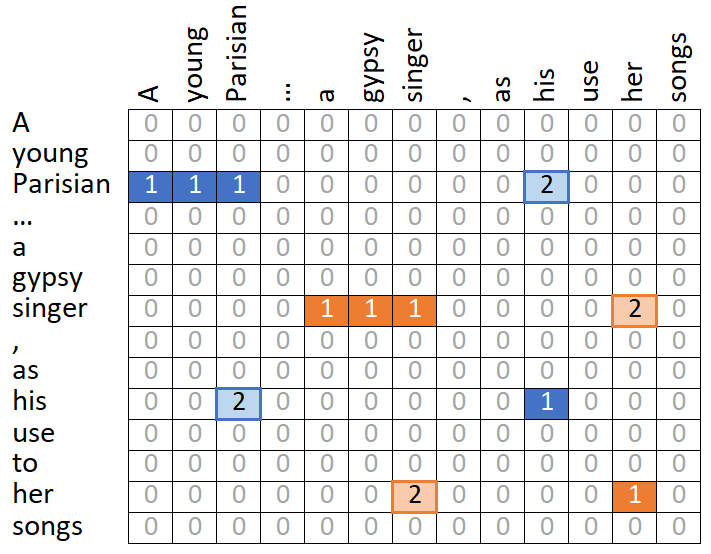}
	\caption{Example of a graph in matrix representation. The connection types are encoded as, 0: no links, 1: mention links, 2: coreference links. }
	\label{fig:coreference:exa-mat}
\end{figure}

We propose a modeling approach that learns coreference at the document-level and takes global decisions. We propose to model mentions and coreference links in a graph structure where the nodes are tokens in the text, and the edges represent the relationships between them. Figures~\ref{fig:coreference:exa} and \ref{fig:coreference:exa-mat} show a short example taken from the CoNLL 2012 dataset \cite{pradhan-etal-2012-conll} showing the graph in two perspectives. Figure~\ref{fig:coreference:exa} shows how the token nodes in a text are connected with edges drawn with arrows. We differentiate the connections between words in a coreference mention, `\emph{mention links}', and the ones among mentions in a cluster, `\emph{coreference links}' (see Sec.~\ref{sec:coreference:graph}). Figure~\ref{fig:coreference:exa-mat} shows the same graph in a matrix representation, where the number in a cell indicates the type of relation between the row and the column. Our model receives a document as input then predicts and iteratively refines the graph of mentions and coreference links.

We follow a similar approach to the Graph-to-Graph Transformer (G2GT) proposed in \cite{mohammadshahi2020recursive, mohammadshahi-henderson-2020-graph} for syntactic parsing, but instead of encoding sentences, we encode documents. Our model predicts the graph in a non-autoregressive manner, then iteratively refines it based on previous predictions.  This recursive process introduces global dependencies between decisions.
Unlike \cite{mohammadshahi2020recursive}, we define different structures for input and output graphs, to reflect the different roles of these graphs. To ensure that locality in the input graph reflects all the relevant relationships, the input graph encodes relations for all mention 
tokens.  This makes the encoding process easier.  
To provide a unique specification of the target graph, the output only encodes a minimal set of connections.  This facilitates prediction.
We initialize the Transformer with pre-trained language models, either BERT \cite{devlin-etal-2019-bert}, or SpanBERT \cite{joshi-etal-2020-spanbert}.

Another difference with \cite{mohammadshahi2020recursive} is that our model predicts two levels of representation.  While they predict the whole graph at each iteration, 
during the first iteration our model only predicts edges that identify mention-spans.  This is because mention detection is a sentence-level phenomenon whose outputs are required as inputs to coreference resolution, which is a discourse-level phenomenon.  But we do not organise these two tasks in a pipeline.  Starting at the second iteration, the model predicts the complete graph.  This allows the model to refine mention decisions given coreference decisions, and vice versa.  In this way, we propose to use iterative graph refinement as an alternative to pipeline architectures for multi-level deep learning models. The iterative process finishes when there are no more changes in the graph or when a maximum number of iterations is reached.

Ideally, the whole document should be encoded at once, but in practice there is a limit on the maximum length. In order to deal with this issue, we propose two strategies: overlapping windows and reduced document. In the first strategy, we split documents into overlapping windows of the maximum allowed size $K$. The segments overlap for a length $K/2$. At decoding time, segments are input in order, and we construct the final graph by joining all graphs from different segments. In the second strategy, we use two networks. The mention-span network is the previously described overlapping model, and we use it for predicting the first graph. For the second network, we reduce the document by including only the tokens of candidate mention-spans, separated by a special token. This network refines the initial graph for the following iterations. 

The experiments show improvements over the relevant baselines and state-of-the-art.  They also indicate that the models reach the best solution in a maximum of three iterations.  Given that we predict the graph at once for each iteration, our model's complexity is lower than the baselines. 
Our contributions are the following:
\begin{itemize}
    \addtolength{\itemsep}{-0.5ex}
	\item We propose a novel modeling approach to coreference resolution using a graph structure and multi-level iterative refinement.
	\item We propose two iterative graph refinement models that can predict the complete entity coreference structure of a document. 
	\item We show improvements over baseline models and the relevant state-of-the-art.
\end{itemize}

The rest of the paper is organized as follows. Section~\ref{sec:coreference:rel} presents a summary of coreference resolution approaches related to this paper.
Section~\ref{sec:coreference:base} briefly describes the fundamentals of state-of-the-art approaches. In Section~\ref{sec:coreference:graph}, we define entity mentions and their coreference links as a graph, and fomulate the task as a sequence-to-graph problem. In Section~\ref{sec:coreference:ite}, we present our iterative refinement solution to global modelling of the coreference graph, and in Section~\ref{sec:coreference:arch}, we present two proposed architectures to address the resulting computational issues. Sections~\ref{sec:coreference:experiments}, \ref{sec:coreference:results} and \ref{sec:coreference:discussion} contain the experimental setup, results and discussion, respectively. Finally, Section~\ref{sec:coreference:conclusion} draws the conclusions of this paper.


\section{Related Work} 
\label{sec:coreference:rel}
The first approaches to coreference resolution (CR) were rule-based systems \cite{lappin-leass-1994-algorithm, manning-etal-2014-stanford}, but eventually, they were outperformed by machine learning approaches \cite{aone-william-1995-evaluating, mccarthy1995using, Mitkov2002} due to annotated corpora's creation. In genral, there are three coreference approaches : mention-pair, entity-mention, and ranking models.
Mention-pair models set coreference as a binary classification problem. 
The initial stage is the mention detection, where the input is raw text, and the output is the locations of each entity mention in the text. Mention detection is done as an independent task in a pipeline model \cite{soon-etal-2001-machine} or as part of an end-to-end model \cite{lee-etal-2017-end}. The next stage is the classification of mention pairs. At first, the best classifiers were decision trees \cite{soon-etal-2001-machine, mccarthy1995using, aone-william-1995-evaluating}, but later, neural networks became the SOTA. The final stage is reconciling the pair-wise decisions to create entity chains, usually by utilizing greedy algorithms or clustering approaches. 
Entity-mention models focus on maintaining single underlying entity representation for each cluster, contrasting the independent pair-wise decisions of mention-pair approaches \cite{clark-manning-2015-entity, clark-manning-2016-improving}. 
Ranking models aim at ranking the possibles antecedent of each mention instead of making binary decisions \cite{wiseman-etal-2016-learning}. An alternative modeling approach is to perform clustering instead of classification \cite{fernandes-etal-2012-latent}. 

SOTA models for CR are mostly based on \citet{lee-etal-2017-end}. They introduced the first end-to-end model that jointly optimizes mention detection and coreference resolution tasks. These neural network-based models also simplify the mention input representation to be word embedding vectors, instead of the traditional pipeline of different linguistic feature extraction tools such as part-of-speech (POS) tagging and dependency parsing. The following models proposed improvements over this work.  \cite{lee-etal-2018-higher} improved the previous model by introducing higher order inference so the entity's mention representation will get iteratively updated with the weighted average of antecedent representations, where the weights are the predictions from the model at the previous iteration. This contrasts with our approach in that we iterate over the whole coreference link graph and we perform discrete decisions at each iteration. \citet{fei-etal-2019-end} use reinforcement learning to directly optimize the model on the evaluation metrics. \citet{joshi-etal-2019-bert} uses BERT embeddings \cite{devlin-etal-2019-bert} as input. \citet{joshi-etal-2020-spanbert} introduced a new SpanBERT embedding model, which is shown to outperform BERT for the CR task. \citet{xu-choi-2020-revealing} showed that higher order inference has low impact on strong models such as SpanBERT.  \citet{toshniwal-etal-2020-learning} proposed a bounded memory model trained to manage limited memory by learning to forget entities. Finally, \citet{wu-etal-2020-corefqa} formulated the problem of coreference resolution as question-answering and trained a model for span prediction. This model has the advantage of being pretrained with larger data-sets from the question-answering task.
 

\section{Baseline: Neural Coreference Resolution} 
\label{sec:coreference:base}

Neural coreference resolution, as formulated in \cite{lee-etal-2017-end, lee-etal-2018-higher}, is a mention-pair approach. It uses an exhaustive method defining mentions as any text span of any size in a document. There, a document $D$  represents a sequence of tokens of size $N$. The objective is to assign an antecedent $y_i$ to each of the $M$ text spans $m_i$ in $D$.  The set of possible antecedents of the span $m_i$ is denoted as  $\mathrm{Y}(i)$. This set contains all text spans with index less than $i$, plus a null antecedent $\epsilon$, $\mathrm{Y}(i) = \{\epsilon, m_1, ...,m_{i-1} \}$. The null antecedent is assigned when:
\begin{enumerate*}[(\alph*)]
	\item the span is not an entity mention,
	\item the span is the first mention of an entity in the document.
\end{enumerate*}
The final mention clusters are constructed greedily by grouping connected spans based on the model predictions during decoding time.

The model is trained to learn a conditional probability distribution over documents $p(y_1, ...,y_n|D)$, assuming independence among each decision of antecedent assignment $y_i$, as follows:
\begin{equation}
	p(y_1, ...,y_M|D) = \prod_{i=1}^M p(y_i|D)
\end{equation}

In \cite{lee-etal-2018-higher}, the probability distribution $p(y_i|D)$ is inferred over $T$ iterations of the model over the same input document. At each iteration $t$, the span representations are updated with the weighted average of all possible antecedents at time $t-1$ where the weights are given by the probability distribution of the model at time $t-1$.  
They called this model high-order coreference resolution since each mention representation considers information from its probable antecedents.  

The training optimization is done using cross-entropy. Given that a mention-span $m_i$ can have more than one true antecedent, the loss considers the sum of probabilities of all true antecedents in the annotated data: 
\begin{equation}\label{eq:coreference:log}
log \prod_{i=1}^M \sum_{y_i \in \mathrm{Y}(i) \cap \mathrm{C}(i)} p(y_i|D)
\end{equation}
where $\mathrm{C}(i)$ indicates the cluster of mention-spans that includes $m_i$ in the annotated data. If the span does not belong to any cluster or all its antecedents have been pruned, then the span is assigned to the null cluster $\mathrm{C}(i)=\{\epsilon\}$. 

This model's complexity is of the order $\mathcal{O}(N^4)$, where $N$ is the document length. The complexity is computed by considering all possible text spans $M$ of the document, so $\mathcal{O}(M) = \mathcal{O}(N^2)$. Then, it considers all possible combinations of span-antecedents $\mathcal{O}(M^2)$. The model prunes spans and candidate antecedents to predetermined maximum numbers in order to maintain computational efficiency.


\section{Graph Modeling} 
\label{sec:coreference:graph}

We propose to model the set of coreference links of a document in a graph structure where the nodes are tokens\footnote{The tokenization of the words in the document, and thus the nodes of the graph, are defined by the input format of the relevant pre-trained Transformer model.} and the edges are links of different types.  Given a document $D=[x_1,...,x_N]$ of size $N$, the coreference graph is defined as the matrix $G \subset \mathbb{N}^{N \times N}$ of links between tokens. Here, the relation type between two tokens, $x_i$ and $x_j$, is encoded with integers and is denoted as $g_{i,j} \in \{0,1,2\}$. We define three relation types: (0) no link, (1) mention link, and (2) coreference link,
as illustrated in Figure~\ref{fig:coreference:exa-mat}. 

\paragraph{Mention links} This type of link serves to identify mentions. 
We define mention links in two different manners depending on whether the graph is an input or output of the model, for functional reasons. When the graph is an input $G^{in}$, there is a directed link from each mention's token to the mention head%
, including the head to itself. 
When the graph is the model's output $G^{out}$, there is only one directed link from the last token of the mention-span to the first token. 
Both encoding methods define a mention-span uniquely, even when having nested mentions; every mention has a unique start-end combination and a unique head. The model utilizes the output for prediction, so it is simpler to predict one single link, whereas, in the input, the model uses links to all tokens to provide a more direct representation of the role of every token in the mention.
	
\paragraph{Mention heads} We simplified the head identification process by considering the first token of a mention span as the head. Although this method is naive, experiments show that this approximation works well enough in practice. However, as some spans can potentially have the same first token in case of nested mentions, we fix this issue by assigning the next token as the head if the first is already the head of any other mention. 
Investigating alternative approaches to mention head identification is future work.
	
\paragraph{Coreference links}  This type of link defines the relationship between a mention and each of its antecedents. We also define coreference links in two different manners depending on whether the graph is an input or output of the model. When the graph is input, there is a link from a mention head token to the head of each mention in the same cluster. 
When the graph is a model's output, the mention should be connected to at least one of its antecedents. If the mention has no antecedent, or corresponds to the first mention of an entity in the text, then it is connected to a null antecedent $\epsilon$. 
We use all possible connections between mentions in an entity cluster for the input so that the model receives a direct input for each coreference relationship. On the other hand, we consider that predicting at least one connection of the mention to its cluster is sufficient to specify the output graph.


The objective is to learn the conditional probability distribution $p(G|D)$. This distribution is initially approximated by assuming independence among each relation $g_{i,j}$ as:
\begin{equation}\label{eq:coreference:p-dist}
p(G|D) = \prod_{i=1}^N \prod_{j=1}^i p(g_{i,j}|D)
\end{equation}
The probability $p(g_{i,j}|D)$ is split in two cases: one for mention links $p_m$ and the other for coreference links $p_c$. The mention link probability is defined as:
\begin{equation}
p_m(g_{i,j}{=}1|D) =  \sigma (W_m \cdot [h_i, h_j])
\end{equation}
where $W_m$ is a parameter matrix, and $h_i$ and $h_j$ are the hidden state representations of the tokens $x_i$ and $x_j$ respectively. This probability indicates whether there is a mention starting at position $j$ and ending at position $i$ of the document $D$. The optimization is done using binary-cross-entropy $loss_m$.

The coreference link probability is defined as:
\begin{equation}
p_c(g_{i,j}{=}2|D) = \frac{exp(W_c \cdot [h_i, h_j])}{\sum_{j' \in \mathrm{A}(i)} exp(W_c \cdot [h_i, h_{j'}])}
\end{equation}
where $W_c$ is a parameter matrix, and $h_i$ and $h_j$ are the hidden state representations of the tokens $x_i$ and $x_j$ respectively. Similar to the baseline, we denote $\mathrm{A}(i)$ as the set of all candidate antecedents of $x_i$. This set contains all mention heads with an index less than $i$, plus a null head $\epsilon$, $\mathrm{A}(i) = \{\epsilon, x_k \; | \; k<i \; \text{and} \; x_k \in \mathrm{H(D)} \} $, and $\mathrm{H(D)}$ is the set of all candidate mention heads in the document. The optimization is done with cross-entropy loss. Given that a mention-span $m_i$ can have more than one true antecedent, the loss considers the sum of probabilities of all true antecedents in the annotated data (as in Equation\eqref{eq:coreference:log}): 
\begin{equation}
loss_c = log \prod_{i \in \mathrm{H}(D)} \sum_{j \in \mathrm{Y}(i) \cap \mathrm{\hat{C}}(i)} p_c(g_{i,j}|D)
\end{equation}
where $\mathrm{\hat{C}}(i)$ indicates the annotated cluster of mention-spans that includes $m_i$ in the annotated data. If the mention does not belong to any cluster, then the span is assigned to the null cluster $\mathrm{\hat{C}}(i)=\{\epsilon\}$. The final loss is the sum of $loss_m$ and $loss_c$.

The token's hidden state representations $\{h_1,..,h_N\}$ are the last hidden layer of a Transformer model. We use various pre-trained Transformer models to initialize the weight parameters, then fine-tune for the coreference task.


\section{Iterative Refinement} 
\label{sec:coreference:ite}

The strong independence assumption made in Equation~\eqref{eq:coreference:p-dist} does not reflect the real scenario and could lead to poor performance. Therefore, we use an iterative refinement approach to model interdependencies between relations, similar to G2GT \cite{mohammadshahi2020recursive}. Under this approach, the model makes $T$ iterations over the same document $D$. At each iteration $t$, the predicted coreference graph $G_t$ is conditioned on the previously predicted one $G_{t-1}$. The model's conditional probability distribution is now defined as follows:
\begin{equation}
p(G^t|D, G^{t-1}) = \prod_{i=1}^N \prod_{j=1}^i p(g_{i,j}|D, G^{t-1})
\end{equation}

This means that the graph should be input to the Transformer model \cite{vaswani2017attention}.  
Following \cite{mohammadshahi2020recursive}, the graph is encoded by inputting an embedding for the type of each relation into the self-attention function of the Transformer :
\begin{multline}
	\mathsf{Attention}(Q, K, V, L_k, L_v) = \cr \textrm{softmax}(\frac{Q \cdot (K+L_k)^{\intercal}}{\sqrt{d}}) \cdot (V + L_v) 
\end{multline}
\vspace{-25pt}
\begin{align}
	\text{where} \quad \quad \quad L_v &= E(G_{t-1}) \cdot W_v  \nonumber \\
	L_k &= E(G_{t-1}) \cdot W_k \nonumber
\end{align}
where $E$ is a matrix of embeddings which encode the types of links in the graph, as illustrated in Figure~\ref{fig:coreference:exa-mat}. Thus, the relationship between a pair of tokens is encoded as an embedding vector which is input when computing the attention function for that pair of tokens. $W_k, W_v$ are weight matrices that serve to specialize $E(G_{t-1})$ to be either \emph{key} or \emph{value} vectors. The complexity of our model is of the order of $\mathcal{O}(N^2 \times T)$, where $N$ is the document length, and $T$ is the number of refinement iterations of the model. 

\begin{figure*}
	\center
	\includegraphics[width=0.65\linewidth]{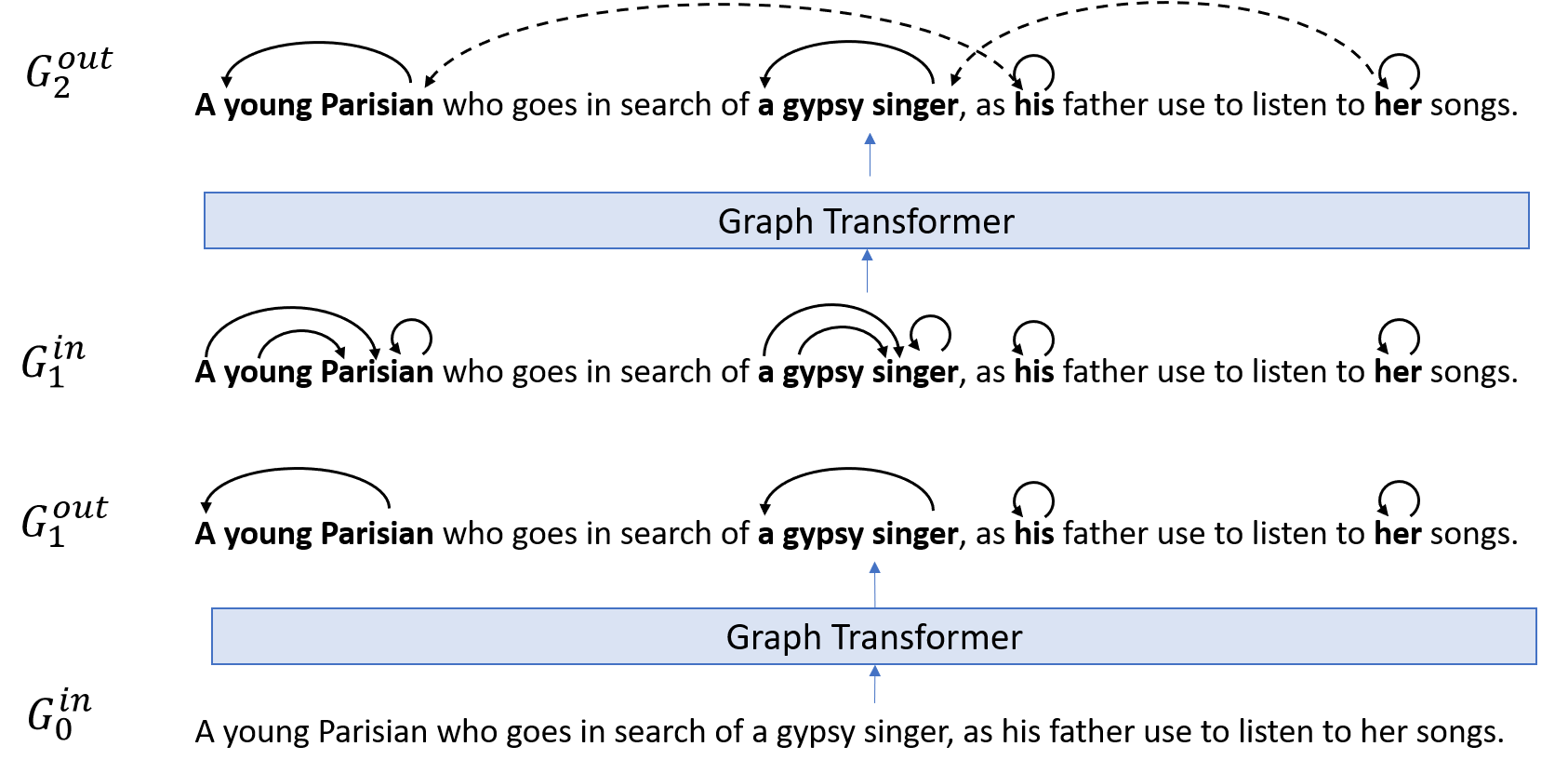}
	\caption{Example of iterations with G2GT.}
	\label{fig:coreference:ite}
\end{figure*}

To illustrate the iterative refinement of a graph, Figure~\ref{fig:coreference:ite} shows an example of two iterations of the model. The mention links are indicated with solid line arrows and the coreference links with dotted arrows. The initial graph matrix $G_0^{in}$ is full of zeros, so no connections are drawn. The first predicted graph $G_1^{out}$ only has mention-links because initially there were no mention heads to be connected. This graph is transformed to serve as input $G_1^{in}$ for the next iteration. Finally, during the second iteration, the model predicts the coreference graph $G_2^{out}$. The model can continue iterating for a maximum of $T$ times.

\begin{figure*}
	\center
	\includegraphics[width=0.65\linewidth]{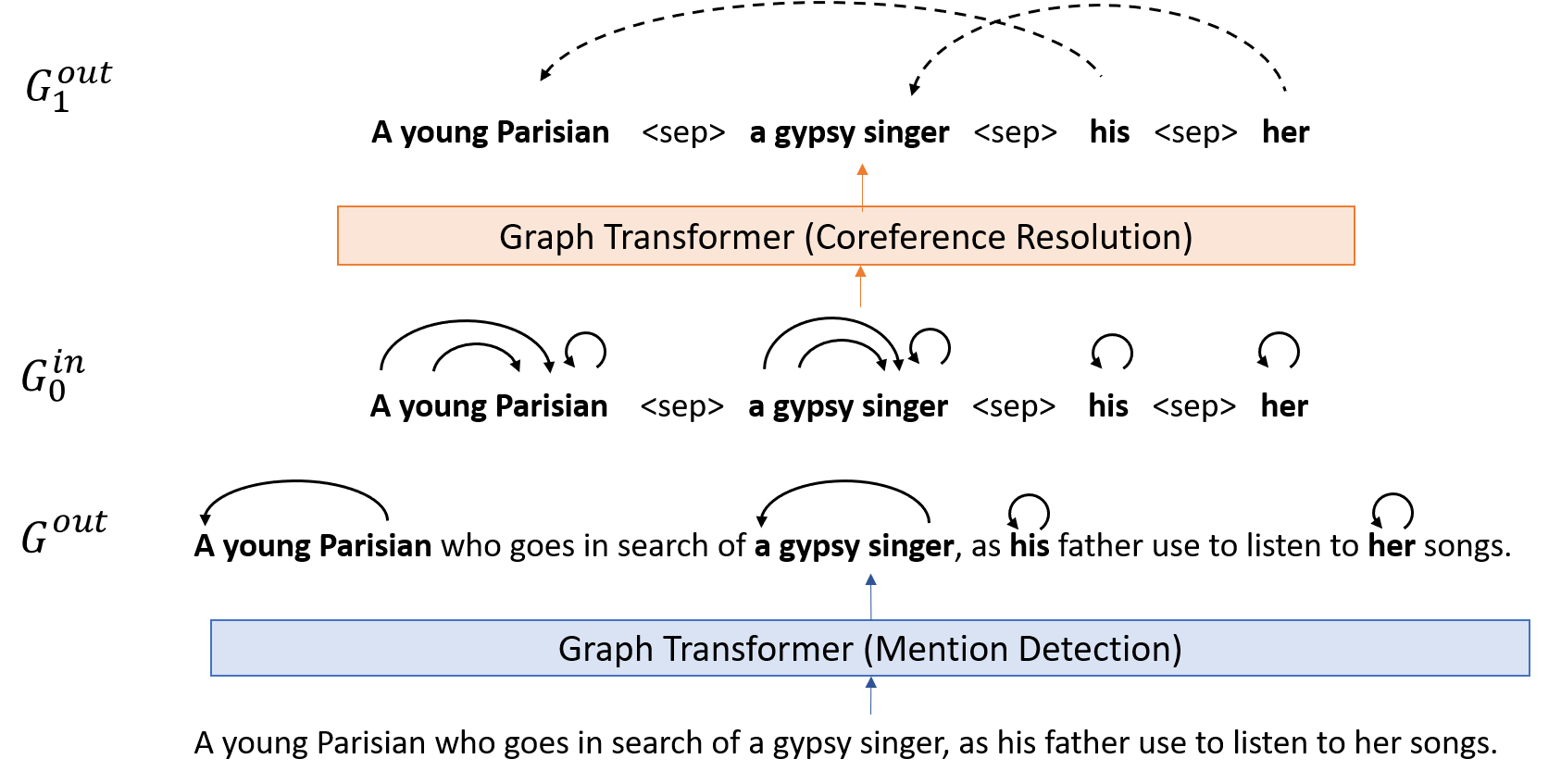}
	\caption{Example of iterations with G2GT in two stages.}
	\label{fig:coreference:ite2}
\end{figure*}


\section{Architectures} 
\label{sec:coreference:arch}

There exists in practice a maximum length for encoding a document due to limited hardware memory. In this section, we describe two strategies to manage this issue: overlapping windows and reduced document. 
In the experiments we also report results for a naive strategy of truncating the documents at the maximum segment length of $K$ for both training and testing.

\subsection{Overlapping Windows}
Here, we split the documents into overlapping segments of the maximum size $K$, with an overlap of $K/2$ tokens. The segments are encoded individually in our G2GT model. During training, each segment is treated as an independent sample. However, during decoding, the segments are decoded in order. The subgraph corresponding to the overlapping part is input to the next segment. 
The union of the segmented graphs forms the final graph.         

\subsection{Reduced Document}
This model has two parts; one to detect mentions and the other to perform coreference resolution. The mention detection is similar to the previously described model. The coreference resolution part receives a shorter version of the document as input. The complete model is described in the following:
\paragraph{Mention Detection} This Transformer is non-iterative so it corresponds to the definition in Equation~\eqref{eq:coreference:p-dist}. To encode the document, we apply overlapping windows, as in the previous section. For prediction, we used the \emph{soft-target} method proposed in \cite{miculicich-henderson-2020-partially}. This method enables the model to increase the recall of detection. 
	Given that the candidate mentions will be fixed for the coreference resolution part, we need to detect most of them here. 
\paragraph{Coreference Resolution} This part is a G2GT with iterative refinement. The input is a shorter version of the document obtained by concatenating the tokens from candidate mention-spans with a separation token in between and removing all other tokens. To maintain coherence in the document, we modify the token input representation to the sum of three vectors:
	\begin{enumerate*}[(\alph*)]
		\item a token embedding,
		\item an embedding of the token's position in the original document, so we retain information of distance between mentions, and
		\item the token's contextualized representation obtained from the mention detection part where the original document is encoded.
	\end{enumerate*}
This second part predicts only coreference links, but the input graph contains both candidate mentions and coreference links.  The set of candidate mentions remains the same across all iterations of this second part, but the mentions are refined in the sense that the final output only includes the mentions which are involved in the final coreference links.

Figure~\ref{fig:coreference:ite2} shows an example of this architecture with one iteration over a document. The mention links are indicated with solid line arrows and the coreference links with dotted arrows. The first model predicts the graph of mention-spans $G^{out}$. This graph is transformed into the input format for the next model $G^{in}_0$. Then, the second model predicts the graph of coreference $G_1^{out}$. Note that this coreference resolution model can continue iterating for $T$ times. The final coreference graph is the output after the final iteration of the second model.  The final set of mentions is only a subset of the mention candidates output by the first model, namely those mentions which participate in coreference links.


\begin{table}
	\centering
	\begin{adjustbox}{width=\linewidth}
	\begin{tabular}{l c c c c} 
		\toprule
		& \textbf{Train} &  \textbf{Dev.} &  \textbf{Test} &  \textbf{Total} \\ \hline			
		\# documents & 2,802 & 343& 348 & 3,493\\
		\# words  & 1.3 M & 160 K& 170 K & 1.6 M\\
		Avg. length  & 464 & 466 & 488 & 458 \\ \hline	
		\# entity changes/clusters & 35 K & 4.5 K & 4.5 K & 44 K   \\
		\# coreference links & 120 K & 14 K & 15 K & 150 K  \\
		\# mentions & 155 K & 19 K & 19 K & 194 K   \\
		\bottomrule
	\end{tabular} 
	\end{adjustbox}
	\caption{Dataset statistics and splits.}
	\label{tab:coreference:data}
\end{table}

%

\section{Experimental Setting} 
\label{sec:coreference:experiments}

\subsection{Dataset}
We use the CoNLL 2012 corpus \cite{pradhan-etal-2012-conll}. It contains data from diverse domains e.g., newswire, magazines, conversations.  We experiment only with the English part. Table~\ref{tab:coreference:data} shows the statistics of the dataset; the average length per document does not exceed 500 words. We pre-process the text to extract sub-word units \cite{sennrich-etal-2016-neural} with BERT tokenizer \cite{wu2016google}. We map the positional annotation of mentions from words to sub-words and retain this mapping for back transformation during evaluation. 

\subsection{Model configuration}
We use the implementation of \citet{wolf2019huggingface}\footnote{\url{https://huggingface.co/transformers/}} of `BERT-base', `BERT-large' \cite{joshi-etal-2019-bert} and `SpanBERT-large' \cite{joshi-etal-2020-spanbert}. All hyper-parameters follow this implementation unless specified otherwise.  

\paragraph{Training}  The G2GT considers an independent loss for each different refinement iteration. There is no back-propagation between refinement iterations because the model makes discrete decisions when predicting the graph for the next refinement step. There are two stopping criteria for the refinement: 
	\begin{enumerate*}[(\alph*)]
		\item when a maximum number of iterations $T$ is reached, or
		\item when there are no more changes in the graph, $G_t = G_{t-1}$. 
	\end{enumerate*}
This criterion  is for both training and testing. 
Our models are trained with a maximum segment length of $K=512$ and a batch size of 1 document. We use Adam \cite{kingma2014adam, wolf2019huggingface} optimizer with a base learning rate of $2e{-}3$ and no warm-up. As our graphs are directed, we use only the lower triangle of $G$ for predictions. The components of the reduced models are trained independently. The coreference resolution follows the currently described training schema. The mention detection model has no iterative refinement step and follows the training schema of the \emph{span scoring soft-target} approach described in \cite{miculicich-henderson-2020-partially}, with $\rho=0.1$. 
	
\paragraph{Evaluation}  At evaluation time, we map back all sub-word units to words and reconstruct the document in CoNLL 2012 format. We use the precision, recall, and F1 score calculated in three different manners: MUC that counts the number of links between mentions, B$^3$ that counts the number of mentions, and CEAF that counts the entity clusters. We did paired bootstrapping re-sampling for significance test following \cite{koehn-2004-statistical}.


\begin{table}[t]
	\centering
	\begin{adjustbox}{width=1\linewidth}
		\begin{tabular}{l l c c c c } 
			\toprule 
			\textbf{Model} & \textbf{Iter.} & \textbf{MUC} & \textbf{$\mathbf{B^3}$} & \textbf{CEAF$_{\phi_4}$} & \textbf{Avg. F1}\\ \hline
			G2GT  & $T=2$ & 75.7 & 68.4 & 65.2 & 69.8 \\
			BERT-base & $T=3$ & 76.9 & 69.3 & 66.0 & 70.7 \\
			\emph{truncated}& $T=4$ & 77.2 & 69.7 & 66.3 & 71.0 \\
			& $T=5$ & 77.2 & 69.7 & 66.3 & 71.0 \\ \hline
			G2GT   & $T=2$ & 80.6 & 69.8 & 67.4 & 72.6 \\ 
			BERT-base & $T=3$ & 81.6	& 71.0 & 68.6 &  73.7 \\
			\emph{overlap}& $T=4$ & 81.5 & 70.9 & 68.7 & 73.7 \\
			& $T=5$ & 81.4 & 70.6 & 68.7 & 73.5 \\\hline
			G2GT   & $T=2$ & 79.2 & 76.1 & 68.5 & 71.6  \\
			BERT-base& $T=3$ & 80.0 & 69.6 & 70.2 & 73.3  \\
			\emph{reduced}& $T=4$ & 81.9 & 70.1 & 71.2 & 74.4 \\
			& $T=5$ & 81.9 & 70.1 & 71.2 & 74.4  \\
			\bottomrule 
		\end{tabular} 
	\end{adjustbox}  
	\caption{Refinement iterations $T$ on the development set (CoNLL 2012).}
	\label{tab:coreference:iter}
	
\end{table}

\begin{table*}[t]
	\centering
	\begin{adjustbox}{width=1\textwidth}
		\begin{tabular}{l c c c | c c c | c c c | c } 
			\toprule \
			& \multicolumn{3}{c}{\textbf{MUC}} & \multicolumn{3}{c}{$\mathbf{B^3}$} & \multicolumn{3}{c}{\textbf{CEAF$_{\phi_4}$}} \\ 
			\textbf{Model} & \textbf{P} & \textbf{R} & \textbf{F1} & \textbf{P} & \textbf{R} & \textbf{F1} & \textbf{P} & \textbf{R} & \textbf{F1} &\textbf{Avg. F1} \\ \hline
			\citet{clark-manning-2015-entity} & 76.1 & 69.4 & 72.6 & 65.6 & 56.0 & 60.4 & 59.4 & 53.0 & 56.0 & 63.0 \\ 
			\citet{wiseman-etal-2016-learning} & 77.5 & 69.8 & 73.4 & 66.8 & 57.0 & 61.5 & 62.1 & 53.9 & 57.7 & 64.2 \\
			\citet{clark-manning-2016-improving} & 79.2 & 70.4 & 74.6 & 69.9 & 58.0 & 63.4 & 63.5 & 55.5 & 59.2 & 65.7 \\
			\citet{lee-etal-2017-end} & 78.4 & 73.4 & 75.8 & 68.6 & 61.8 & 65.0 & 62.7 & 59.0 & 60.8 &  67.2  \\ 
			\citet{fei-etal-2019-end} & 85.4 & 77.9 & 81.4 & 77.9 & 66.4 & 71.7 & 70.6 & 66.3 & 68.4  & 73.8 \\ 
			\citet{xu-choi-2020-revealing} & \textbf{85.9} & 85.5 &85.7 &79.0 &78.9 &79.0 &\textbf{76.7}& 75.2& 75.9& 80.2 \\
			\citet{wu-etal-2020-corefqa} & {\color{gray}88.6} & {\color{gray}87.4} & {\color{gray}88.0} & {\color{gray}82.4}& {\color{gray}82.0} & {\color{gray}82.2} & {\color{gray}79.9} & {\color{gray}78.3} & {\color{gray}79.1}& {\color{gray}83.1} \\
			\hline
			Baseline \cite{lee-etal-2018-higher}   & 81.4 & 79.5 & 80.4 & 72.2 & 69.5 & 70.8 & 68.2 & 67.1 & 67.6 & 73.0 \\
			+ BERT-base \cite{joshi-etal-2019-bert}  & 80.4 & 82.3 & 81.4 & 69.6 & 73.8 & 71.7 & 69.0 & 68.5 & 68.8 & 73.9 \\
			+ BERT-large \cite{joshi-etal-2019-bert}  & 84.7 & 82.4 & 83.5 & 76.5 & 74.0 & 75.3 & 74.1 & 69.8 & 71.9 & 76.9 \\  
			+ SpanBERT-large \cite{joshi-etal-2020-spanbert} & 85.8 & 84.8 &85.3 &78.3 &77.9 &78.1& 76.4& 74.2 &75.3 &79.6 \\ \hline
			G2GT BERT-base \emph{truncated} & 78.4 & 77.9 & 78.1 & 69.6 & 71.0 & 70.3 & 66.8 & 67.3 & 67.0 & 71.8 \\
			G2GT BERT-base \emph{overlap} & 81.2 & 82.8 & 82.0 & 69.8 & 73.6 & 71.6 & 69.6 & 69.3 & 69.4 & 74.4 \\
			G2GT BERT-base \emph{reduced} & 83.4 & 83.1 & 83.2 & 70.1 & 73.7 & 71.9 & 72.1 & 70.1 & 71.0 & 75.4 \\ 
			G2GT BERT-large \emph{truncated} & 80.1 & 79.2 & 79.6 & 71.3 & 71.0 & 71.1 & 69.1 & 68.8 & 68.9 & 73.2 \\
			G2GT BERT-large \emph{overlap} & 83.5 & 83.2 & 83.3 & 74.5 & 74.1 & 74.3 & 75.2 & 70.1 & 72.6 & 76.7 \\
			G2GT BERT-large \emph{reduced} & 84.7 & 83.1 & 83.9 & 76.8 & 74.0 & 75.4 & 75.3 & 70.1 & 72.6 & 77.3 \\
			G2GT SpanBERT-large \emph{overlap} & 85.8 & 84.9 &85.3 & 78.7 &	78.0& 78.3 & 76.4 & 74.5 & 75.4 & 79.7 \\
			G2GT SpanBERT-large \emph{reduced} & \textbf{85.9}	&\textbf{86.0}$^{*\dagger}$&	\textbf{85.9}$^{*}$&	\textbf{79.3}$^{*}$&	\textbf{79.4}$^{*\dagger}$&	\textbf{79.3}$^{*}$&	76.4&	\textbf{75.9}$^{*}$&	\textbf{76.1}$^{*}$&	\textbf{80.5}$^{*}$ \\ 
			\bottomrule 
		\end{tabular} 
	\end{adjustbox}  
	\caption{Evaluation on the test set (CoNLL 2012).~  $*$ significant at p < 0.01 compared to \citep{joshi-etal-2020-spanbert},~ $\dagger$ significant at p < 0.05 compared to \citep{xu-choi-2020-revealing}}.
	\label{tab:coreference:results}
	
\end{table*}

\section{Results Analysis} 
\label{sec:coreference:results}
This section describes the results of various baselines and our models. 
First, we analyze the optimum number of refinement iterations, and then we show results using the best models.

Table~\ref{tab:coreference:iter} shows the performance of our G2GT models when varying the maximum number of refinement iterations $T$ from 2 to 5 ($T{=}1$ is mention detection only). The results are in terms of the F1 score of the three coreference metrics and the average. All three implementations shown in the table perform the best when using $T{=}4$. There is a significant decrease in performance when the graphs are not refined, $T{=}2$, showing the importance of modelling the interdependencies between coreference relations. 

Table~\ref{tab:coreference:results} shows the evaluation results on the test set in terms of precision (P), recall (R), and F1 score for each metric. The last column displays the average F1 of the three metrics. The first section of the table exhibits scores of different coreference resolution systems from the literature.  The second section shows the result of the `Baseline' \cite{lee-etal-2018-higher} system described in Section~\ref{sec:coreference:base}. This model uses ELMo
\cite{peters-etal-2018-deep} instead of BERT to obtain token representations. Baseline plus `BERT-base', `BERT-large' \cite{joshi-etal-2019-bert} and ` SpanBERT-large' \cite{joshi-etal-2020-spanbert} correspond to the baseline using those pretrained representations. We copy all these values from the original papers. The last section of the table presents scores of our graph-to-graph  models with iterative refinement. `\emph{truncated}' is our model with no special treatment for document length; the documents are truncated at the maximum segment length of $K$. `\emph{overlap}' and `\emph{reduce}' are the models described in Section~\ref{sec:coreference:arch}.

As expected, pre-training with SpanBERT results in better scores than with BERT, and BERT-large is better than BERT-base.
Not surprisingly, `G2GT BERT-base truncated' and `G2GT BERT-large truncated' perform poorly in comparison to the baseline because their information is incomplete. For BERT-base, both the `\emph{overlap}' and `\emph{reduce}' models have better scores than the comparable baseline.  For BERT-large and SpanBERT, the `\emph{overlap}' model has similar scores to the baseline, but the `\emph{reduce}' model consistently improves over the baseline. 

Preliminary experiments with G2GT `\emph{overlap}' in a pipeline approach, where mention detection is performed before coreference, showed that it is not better than in a joint approach showed here. Overall, our G2GT `\emph{reduce}' method consistently shows the highest scores across all the models for each pre-trained model. 
Our models do not surpass SOTA \cite{wu-etal-2020-corefqa} (shown in grey), but as mentioned before, this SOTA model is also trained on the much more abundant data from the question-answering task, and so it is not directly comparable to our model. We leave the issue of incorporating additional data into the training of our model to future work.

\section{Discussion} 
\label{sec:coreference:discussion}

These results support our claim that coreference resolution benefits from making global coreference decisions using document-level information.  First, refinement of coreference decisions using global information about other coreference decisions clearly improves accuracy, as indicated by the improved scores for models with more than one coreference iteration in Table~\ref{tab:coreference:iter}.  
Second, the model which is able to combine information from the entire document, G2GT `\emph{reduce}', is clearly better than the model which performs the task on large windows of text and then merges the results, G2GT `\emph{overlap}'.
We believe that the benefits of full-document iterative refinement will extend to other discourse-level phenomena, and that the G2GT architecture will be an effective way to achieve this benefit.

One issue with our method is the necessity to iteratively pass the input through an expensive encoder model more than once.
However, the number of iterations needed is small, and results in  significant improvement. 

The length management methods would not be necessary if we had more efficient pre-trained Transformer models or larger-memory GPU hardware which could handle longer sequences.  However, the computational cost of very large Transformers will always be an issue, so in general there is a need to address the issue of how to reduce the number of inputs when modelling phenomena which require large contexts, such as coreference resolution.  This paper contributes towards addressing this general issue.

\section{Conclusion} 
\label{sec:coreference:conclusion}

We proposed a G2GT model with iterative refinement for coreference resolution. For this purpose, we define a graph structure to encode coreference links contained in a document. That enables our model to predict the complete coreference graph at once. The graph is then refined in a recursive manner, iterating the model conditioned on the document and the graph prediction from the previous step.
This allows global modelling of all coreference decisions using all document-level information, but it introduces computational issues for longer documents.
We experimented with two methods to manage long documents and maintain computational efficiency. The first method encodes the document in overlapping segments. The second method reduces the set of tokens which are input. 

The evaluation shows that both methods can outperform a comparable baseline, and that the second method has better performance than the first one and than all other comparable models. This experiment shows that global decisions and document-level information are useful to improve coreference and thus should not be ignored. It also shows that the models can benefit from increasingly powerful pre-trained language models, BERT-base \cite{devlin-etal-2019-bert}, BERT-large \cite{devlin-etal-2019-bert}, and SpanBERT \cite{joshi-etal-2020-spanbert}. 

By empirically showing the benefits of making global decisions and using document-level information in coreference resolution, this work motivates further work on this topic. In addition, the model designs developed in this work provide a viable approach to addressing the related issues. Addressing the computational issues with modelling large documents in Transformers is an area of active research, and our proposed methods could be improved in future work.

\section*{Acknowledgements}

This work was supported in part by the Swiss National Science Foundation, under grants 200021\_178862 and CRSII5\_180320.

\bibliography{anthology,custom}
\bibliographystyle{acl_natbib}

\end{document}